\documentclass[runningheads]{llncs}
\usepackage[T1]{fontenc}
\usepackage{graphicx}

\usepackage{comment}
\usepackage{tikz}
\usetikzlibrary{shapes,decorations,arrows,calc,arrows.meta,fit,positioning}
\tikzset{
    -Latex,auto,node distance =1 cm and 1 cm,semithick,
    state/.style ={ellipse, draw, minimum width = 0.7 cm},
    point/.style = {circle, draw, inner sep=0.04cm,fill,node contents={}},
    bidirected/.style={Latex-Latex,dashed},
    el/.style = {inner sep=2pt, align=left, sloped}
}
\usepackage{changepage}
\begin{document}
\title{What The F*ck Is Artificial General Intelligence?}
%
%
\author{Michael Timothy Bennett\inst{1}\orcidID{0000-0001-6895-8782}}
\authorrunning{Michael Timothy Bennett}
%
\institute{The Australian National University\\
\email{michael.bennett@anu.edu.au}}
\maketitle              
\begin{abstract}
Artificial general intelligence (AGI) is an established field of research. Yet some have questioned if the term still has meaning. AGI has been subject to so much hype and speculation it has become something of a Rorschach test. Melanie Mitchell argues the debate will only be settled through long term, scientific investigation. To that end here is a short, accessible and provocative overview of AGI. I compare definitions of intelligence, settling on intelligence in terms of adaptation and AGI as an artificial scientist. Taking my cue from Sutton's Bitter Lesson I describe two foundational tools used to build adaptive systems: search and approximation. I compare pros, cons, hybrids and architectures like o3, AlphaGo, AERA, NARS and Hyperon. I then discuss overall meta-approaches to making systems behave more intelligently. I divide them into scale-maxing, simp-maxing, w-maxing based on the Bitter Lesson, Ockham's and Bennett's Razors. These maximise resources, simplicity of form, and the weakness of constraints on functionality. I discuss examples including AIXI, the free energy principle and The Embiggening of language models. I conclude that though scale-maxed approximation dominates, AGI will be a fusion of tools and meta-approaches. The Embiggening was enabled by improvements in hardware. Now the bottlenecks are sample and energy efficiency.

\end{abstract}

\section{Introduction}

Picture a machine endowed with human intellect. In its most simplistic form, that is Artificial General Intelligence (AGI) \cite{russell2022}. AGI is also a well established and rigorous field of research \cite{goertzel2014}.
However public perception of AGI is plagued by wild speculation and hype. 
Some see it as Skynet waiting to pounce \cite{bostrom2014superintelligence,russell2022}.
Others, like Melanie Mitchell, question if the term still has any meaning \cite{mitchell2024}. Speculation and hype have reduced it to a Rorschach test. As Mitchell points out, the debate will not be settled not by media but by rigorous, scientific research. Here I present a short and accessible survey to that end. It is framed in intentionally provocative terms, to spark debate\footnote{There is precedent for the use of profanity in a paper title \cite{krauth2012}. However this paper will be published in the 2025 AGI proceedings under the title ``What Is Artificial General Intelligence'' because Anton threw a tantrum. The real name of the paper remains What the F*ck Is Artificial General Intelligence. Please cite it as that. I'd like to dedicate this footnote to Anton's pearl clutching. Good job Anton.}.
\section{Intelligence}
I'll begin by defining intelligence and AGI. There are a number of positions \cite{wang2019,goertzel2014,goertzel2023generativeaivsagi,thorisson2012,wang2006rigid,legg2007,chollet2019,bennett2024a}. Some peg AGI to \textbf{human-level performance} across a broad range of tasks \cite{sternberg1984,russell2022}. This is is intuitive, but anthropocentric and hard to quantify\footnote{More accurately, I might say it is too easy to quantify. There are so many ways we might quantify it. It is ambiguous and too weak a criteria to be much use for anything but padding out Sam Altman's Twitter feed.}. Chollet argues intelligence is a measure of the ability to generalise and \textbf{acquire new skills}. He argues AGI can do this at least as well as a human \cite{chollet2019}. He attempts to quantify the ability to acquire new skills, which can encompass the aforementioned anthropocentric definition. His formalism resembles Legg-Hutter intelligence. Legg and Hutter argued intelligence is an ability to \textbf{satisfy goals in a wide range of environments} \cite{legg2007}\footnote{This treats intelligence as implicitly separable from goals, endorsing the orthogonality thesis \cite{bostrom2012}.}. Chollet's definition descends from Legg-Hutter. It is based on Ockham's Razor. They both use Kolmogorov complexity. They both equate simplicity with generality. They both seek to quantify intelligence, and they are both highly subjective because they treat intelligence as a property of software interacting with the world through an interpreter \cite{orseau2012,orseau2012b,leike2015,bennett2024b}. 

That a problem. Why? Because if I develop an AI for some purpose, then \textit{I} decide whether it has fulfilled that purpose, and \textit{I} am part of the agent's environment. The environment is where \textit{objective} success or failure is decided. Assume $\mathcal{C}$ is a space of software programs, and $\Gamma$ is a space of behaviours. Imagine $f_1 \in \mathcal{C}$ is AI software, $f_2 : \mathcal{C} \rightarrow \Gamma$ is the hardware on which it runs, and $f_3 : \Gamma \rightarrow \{0,1\}$ is the environment (including me) where success is decided. Success is a matter of $f_3(f_2(f_1))$. The behaviour of $f_3(f_2(f_1))$ can be changed by changing $f_2$ or $f_3$ \cite{bennett2024a}. It is pointless to make claims about $f_3(f_2(f_1))$ based on $f_1$ alone. $f_1$ and $f_2$ are like mind and body. Every choice of embodiment biases the system in some way. Each movement it makes constrains the space of possibilities, much like a constraint expressed in a formal language. Complexity is a property of how a body interprets information \cite{bennett2024b}. The choice of Universal Turing Machine can make any software agent optimal according to Legg-Hutter intelligence \cite{leike2015}.

The idea of AI as a software mind is called \textbf{computational dualism} \cite{bennett2024a}\footnote{Computational dualism is grounded in lengthy formal definitions and derivations given elsewhere \cite{bennett2025f,bennett2024b,bennett2024a,bennett2022b,bennett2025a}. For this survey that level of formality would be counterproductive.}. It is a reference to the work of Descartes, who in 1637 argued the pineal gland mediates between mind and body. AI researchers have exchanged the pineal gland for a Turing machine. So what is the alternative? Wang defines intelligence as \textbf{adaptation with limited resources} \cite{wang2019}. This leaves room for us to avoid dualism, and it implies the ability to satisfy goals in a wide range of environments anyway \cite{bennett2024a}. 

An attempt was made to resolve computational dualism and formalise intelligence as objective adaptability. It does so by formalising software, hardware and environment together \cite{bennett2024a}. It formalises intelligence as a measure of the \textbf{ability to complete a wide range of tasks} \cite{bennett2025a}. This dispenses with the separation of goals and intelligence in favour of a whole-of-system model that treats the purpose of a system as what it does. One's body implies a set of goals and subgoals. Body, environment and goals together form a task, by which I mean a purpose and a means of fulfilling it. If $A$ completes a superset of tasks that $B$ completes, then $A$ is more adaptable than $B$. This encompasses both \textbf{sample and energy efficiency}. It is how fast a system can adapt and how much energy it needs to do so. This is the definition I will use for this survey. I'll consider an AGI to be a system that adapts at least as generally as a human scientist \cite{bennettmaruyama2022b}. An \textbf{artificial scientist} can prioritise, plan and perform useful experiments. This requires autonomy, agency, motives, an ability to learn cause and effect and the ability to balance exploring to acquire knowledge with acting to profit from it \cite{goertzel2021,wang2006rigid,thorisson2012,thorisson2014autonomousNatCom,sutton2018}.

Artificial intelligence (AI) and machine learning (ML) are typically divided up into buckets like supervised learning, reinforcement learning, regression, classification, planning and so on. These are not useful categories for AGI, because an artificial scientist must be able to do all of these things. Instead, I will take my cue from Sutton's Bitter Lesson. It acknowledges that generally applicable tools can be used to learn any behaviour \cite{sutton2019}, if we scale up resources (compute, memory, data etc).

\section{Tools}
\paragraph{Search:} Informally, by search I mean systems that take structure, and then \textit{construct} a solution within the confines of that structure. For example, take a map and then plan a route by trying first every combination of one, then two, then three and more turns until you find the smallest sequence of turns that end at your destination. 
Search typically refers to algorithms like A* used for symbolic reasoning and planning problems \cite{russellandnorvig2021}. These involve describing a problem and goal as a set of rules, and then constructing possible courses of action until one is found that obeys all the rules. Heurstics are used to construct a solution faster\footnote{An example of a heuristic is a function that takes a sequence of turns and tells you how far the end is from your destination.}. In theory any problem can be framed as a search problem\footnote{At first glance it might be tempting to object, and say search can only be applied for a well-defined problem with unambiguous rules and goals. However search can easily be applied to poorly defined problems. In the absence of well defined rules and goals, search can be used to infer rules and goals from observed data. It can do this by treating observed data or subsets thereof as rules, and searching the space of all possible criteria until one is found that explains some or all of observed data\cite{bennett2023b,bennett2025e}. Everything can be reduced to a search problem for much the same reason that every imperative program (instruction like ``do this'') can be translated into an equivalent declarative program (an assertion like ``the pen is red'')\cite{howard1980}; it is simply a matter of framing.}. 
Search has \textbf{advantages}. It produces verifiably correct and interpretable answers. It excels at planning \cite{kautz92} and is typically used in map software. It can prove theorems \cite{newell1956}. In the 90s, search defeated the world chess champion \cite{campbell2002}. Search can be used to learn, by iterating through possible hypotheses or models until one is found that conforms to observed data.
However it also has \textbf{disadvantages}. Iterating through large state spaces is expensive. Hand crafted constraints can be added to reduce the search space, but that is not very scalable.
Search tends to be sequential\footnote{Tends to in present day implementations. Doesn't need to be.}, making it ill suited to take full advantage modern hardware, which was originally designed to parallelise graphical rendering and physics simulation in games. Only later was this hardware adapted for AI \cite{kirk2007}. Parallel search algorithms exist but there is a lot of room for improvement \cite{rossi2006_csp,edelkamp2012_distSearch,zhou2015_parallelAstar,oswald2023}. The consequence is that search is only really practical at a higher levels of abstraction, where problems are represented using a small number of abstract symbols or well defined parts.  

\paragraph{Approximation:}
Sutton's Bitter Lesson described the alternative to search as learning. However search can be used to learn \cite{bennett2023b}, so to avoid confusion I'll use the term approximation instead. In any case most of modern machine learning is approximate. Typically it involves taking a model that can map inputs to outputs, then changing its parameters to so that the relation between inputs and outputs approximates training data. For example, convolutional neural networks can be taught to classify the contents of images \cite{krizhevsky2017}. Transformers trained on large corpus of text can generate human-like responses \cite{vaswani2017}. Approximation is imprecise, but for that very reason it is great at dealing with noisy data and large state spaces. It is easy to parallelise and scale on current hardware. 
There are drawbacks. Approximation is unreliable\footnote{Imprecision implies a sort of unreliability.}, because it is by definition only approximate. It is not easily interpretable \cite{ribeiro2016}. Most importantly, current methods are extremely sample and energy inefficient \cite{strubell2019energy,bennett2025a}. This makes them less adaptable. Sample inefficiency doesn't just mean a model is slower to learn. It means the model does not cope well with anything outside the norm. In layman's terms, an approximation is mid. If two models are trained on the same amount of data, then the more sample efficient model will deal as well or better with edge cases\footnote{The proofs are given elsewhere \cite{bennett2023b,bennett2025e,bennett2025d,bennett2025f}, but suppose for a second I learn faster than you (more sample efficient). We both learn how to fix a table from the same three examples of someone fixing a table. We can now both fix the table with 100\% accuracy. However table fixing is an example of fixing in general. If I am more sample efficient, I must now be able to fix something you cannot. An edge case, like a chair. Now, there's an upper bound for when we have both learned everything, but given that is impossible given finite resources it will always be the case that the more sample efficient learner will deal better with edge cases, all else being equal.}.

\paragraph{Hybrids and Architectures:}
Hybrids are those systems which don't fit neatly into search or approximation. For example collectives of living cells self-organise and adapt. They can traverse a morphospace during development or regeneration \cite{levin2024}, which is like search. Animals mimic and thus approximate behaviour. It is difficult to argue biolgical self-organisation falls neatly into either search or approximation. Also, our current methods for search and approximation have complementary strengths and weaknesses. They can be combined to get the best of both worlds.
Hybrids are inherently more general because they're not tied to one playbook. Need precision? Search. Got a mess of unstructured data? Approximate. By fusing their strengths, hybrids promise robustness where single-track systems choke \cite{bennettmaruyama2022a}.  

Perhaps the simplest example of a hybrid is \textbf{AlphaGo} \cite{silver2016}. It vanquished Go's world champion using a combination of search and approximation. Search enabled AlphaGo to explore potential sequences of moves within the game's constraints. Deep neural networks then approximated how likely sequences were to win. Intuitively, think of these as `how to play' and `how to win' respectively. This synergy allowed AlphaGo to surpass human champions, demonstrating the potential of hybrid approaches in mastering complex, strategic tasks.

Search tends to be applied in the context of high level symbolic abstractions that depend on human interpretation. For example, the word `cat' is just a sound until someone interprets it. It must be decided why and how a particular problem is represented using a particular language or set of symbols. This is the symbol grounding problem \cite{harnad1990}. \textbf{Neuro-symbolic hybrids} attempt to address it \cite{garcez2019}. These systems typically employ neural networks to interpret raw input, converting them into symbolic representations that encapsulate meaning. Search can then be applied to these representations to enable tasks such as planning or logical inference. However it should be noted that the complexity of a problem depends on how it is represented \cite{bennett2024b}, and not all choices of symbolic representation are equal. 
Another hybrid approach is \textbf{structured reinforcement learning}. It leverages approximation to reduce high-dimensional raw sensory data to a more manageable symbolic format. Convolutional autoencoders are used to compress high-dimensional data into concise, symbolic forms that try to capture what is \textit{relevant} in the input, enabling more effective adaptation to dynamic environments \cite{garnelo2016deepsymbolicreinforcementlearning}.
More recent examples include OpenAI's o3 and DeepMind's AlphaGeometry. \textbf{o3} employs chain-of-thought reasoning, blending approximation with a structured processes for complex problem-solving \cite{openai2025o3}. \textbf{AlphaGeometry} combines neural networks with symbolic reasoning to solve geometry problems \cite{trinh2024}. These systems exemplify the shift towards hybrid approaches for more capable AI.

Finally, there are comprehensive frameworks designed to be generally intelligent. Cognitive architectures and autonomous machines constructed from modules that each serve a different purpose. Perception, memory, and reasoning modules. System 1 and system 2. For example scaffolding can be applied to neural networks to facilitate persistent identity and memory \cite{perrier2025positionstopactinglike}. Pioneering examples include cognitive architectures like SOAR \cite{laird2012soar} and ACT-R \cite{Anderson2004}. More recent examples include Hyperon, Autocatalytic Endogenous Reflective Architecture (AERA) and the Non-Axiomatic Reasoning System (NARS). 

\begin{itemize}
    \item \textbf{Hyperon} is a modular, distributed system integrating probabilistic logic networks, neural networks, and a knowledge metagraph for holistic cognition \cite{goertzel2021,goertzel2023,goertzel2023generativeaivsagi}. It is highly distributed, modular, scalable and self-organising. This makes it a versatile AGI platform that can integrate new technology as it develops. For example it appears Hyperon will soon incorporate a discrete form of active inference \cite{friston2010,goertzel2024}. 
    \item \textbf{AERA} self-programs, reflecting on its own symbolic structures while learning statistically. It emphasises analogy, causality, autonomy and growth, with predictive modelling supporting proactive adaptation \cite{nivel2013,thorisson2012,thorisson2020,sheikhlar2024,eberding2024}. 
    \item \textbf{NARS} rejects rigid axioms for a fluid, adaptive logic. Operating under the Assumption of Insufficient Knowledge and Resources (AIKR), NARS reasons with incomplete, uncertain data via a non-axiomatic framework. It integrates symbolic reasoning with probabilistic inference, using a custom inheritance-based logic (NAL) to derive conclusions from limited evidence. Designed for real-time adaptability, NARS learns incrementally, refining its knowledge base as new inputs arrive \cite{wang2006rigid,hammer2020}.
\end{itemize}
Hybrid systems have obvious advantages. They can be more efficient, interpretable, they can integrate human priors effectively and above all they allow for autonomy. It can also be difficult to harmonise disparate methodologies. A lack of robust theoretical guidance risks hybrids being ad hoc rather than principled. This brings us to the final piece of the puzzle.

\clearpage
\section{Meta-Approaches}
If we're to build an artificial scientist, we need a clear idea of what we're optimising for. What constitutes a `good' hypothesis? We need a theory that predicts whether one model adapts better than another. There are many such cases. These aren't algorithms so much as they are philosophies with teeth. Guiding principles like `always choose the least specific solution' or `delegate control instead of micromanaging tasks'. I call them meta-approaches.
Examples include the Free Energy Principle \cite{friston2010}, Universal Artificial Intelligence (UAI) \cite{hutter2024introduction}, the Minimum Description Length Principle \cite{rissanen1978}, the scaling hypothesis \cite{sutton2019}, Stack Theory or Pancomputational Enactivism \cite{bennett2024a,bennett2025a} and even organisational principles like the military doctrine of Mission Command \cite{ingesson2016}. 
To simplify matters I put meta-approaches into buckets based on what they share in common. Scale-maxing, simp-maxing, and w-maxing. These maximise resources, simplicity of form and versatility of function respectively. They can be understood in terms of Sutton's Bitter Lesson, Ockham's Razor and Bennett's Razor respectively \cite{sutton2019,sober2015,bennett2023b}. I'll begin with the elephant in the room. 

\paragraph{Scale-Maxed:}
As Sutton observed we seem to be able to just crank up compute, data and model size and get something that looks like intelligence. Scale-maxed approximation has defined recent history. I call this period `The Embiggening'. Language and vision models just got bigger as a bottleneck in compute was removed by technology originally developed for games \cite{kirk2007}. GPT-3? 175 billion parameters, 45TB of text, and it's churning out essays, code, and creepy love letters \cite{roose2023conversation}. AlphaFold 2? Threw a data center at protein folding and cracked a puzzle that had biologists weeping for decades \cite{jumper2021}. 
However performance gains diminish with scale \cite{kaplan2020scaling}. The energy bill is a nightmare \cite{strubell2019energy}. Worst of all is the sample inefficiency. Today, scale-maxed approximations like GPT-4 struggle with novelty and always will, because novelty is by definition that of which we have few examples. Confront a large language model (LLM) with the genuinely unusual, and it'll flail like a toddler in a calculus exam. The Bitter Lesson says scale will eventually work. It will eventually identify all other meta-approaches. Fine. Scale will eventually work, but \textit{eventually} is doing all the work in that sentence. 

\paragraph{Simp-Maxed:}
Simplicity maximisation (simp-maxing) assumes the most accurate predictions are made by the simplest models. Simpler models can be written as shorter programs, so AI researchers have for a long time equated intelligence with compression \cite{chaitin1966}. There are many such cases. Regularization is the most common (e.g. dropout \cite{srivastava2014}). Likewise the Minimum Description Length principle (MDL) lends itself well to selecting hypotheses at high levels of symbolic abstraction \cite{rissanen1978}. 

Then there is UAI. The AIXI UAI is a mathematical formalism of superintelligence \cite{hutter2024introduction}. It equates simplicity with compressibility, and bases its decisions on the most compressed representations of its history. The length of such an smallest self-extracting archive of a dataset is the Kolmogorov Complexity of the dataset \cite{kolmogorov1963,kolmogorov1968}. Solomonoff induction assigns probabilities to programs based on Kolmogorov complexity \cite{solomonoff_1964a,solomonoff_1964b}\footnote{Now available for Boolean circuits \cite{wyeth2023}.}. AIXI uses Solomonoff induction to identify the best models of its environment\footnote{Now available for incomputable environments \cite{oswald2024extension}.}. It can then choose the best possible actions based on those models. Conversely, just as we can say the optimal agent is the one that identifies the simplest models, we can use this Kolmogorov Complexity to measure intelligence\footnote{Oh sure, Kolmogorov Complexity is incomputable. Nobody cares. I can approximate it arbitrarily.}. We can check to see if an agent learns one of the simplest models, or how close to simplest its model might be. That is Legg-Hutter intelligence: a measure of intelligence \cite{legg2007}. Chollet's measure, mentioned in the introduction, is similarly based on Kolmogorov Complexity \cite{chollet2019}. Unfortunately Kolmogorov Complexity is incomputable, but working \textit{approximations} of both AIXI and Legg-Hutter intelligence exist \cite{hutter2007,legg2011}. These represent the universal upper bounds on intelligence.

Except they don't. AIXI is a case of computational dualism \cite{bennett2024a}. Complexity is a property of form, not function. Kolmogorov complexity hinges on your choice of Turing machine \cite{leike2015}\footnote{A choice of Turing Machine is a choice of interpreter is a choice of descriptive language.}. In an interactive setting, there is an interpreter between the software mind and the world it inhabits. Complexity need not have any bearing on reality at all. However, there is a correlation between simplicity of form and generalisation of function. There are reasons for this correlation \cite{bennett2024b}. First, a bounded system can contain only a finite amount of information \cite{bekenstein1981}. Second, a goal-directed process like natural selection can select for systems that make accurate predictions. Third, to make accurate predictions using a finite vocabulary of representations of varying complexity, simpler forms must express more generalisable, \textit{weaker} constraints on functionality \cite{bennett2024b}, which brings us to our third meta-approach.

\paragraph{W-Maxed:}
Computational dualism frames intelligence as a disembodied software policy interacting with the world through an interpreter. The alternative is cognition as a process taking place within the environment, as a part of the environment \cite{dreyfus1972}. This is called \textit{enactive} cognition \cite{thompson2007,vervaeke2012}. A formalisation of enactive cognition must formalise the system as a whole, not parts. This is a challenge. One formalism, does this by moving the problem of interpretation outside the environment \cite{bennett2024a}, basically saying ``whatever determines the laws of physics is the interpreter'' and assuming it is unknowable\footnote{This is called Stack Theory, because it frames everything as an infinite stack of abstraction layers and assumes the bottom layer is unkowable or does not exist\cite{bennett2025f}.}. This is useful to examine not just complexity of form and generalisation of function, but the relationship between the two in all possible environments. It was subsequently shown that generalisation stems from weakening the constraints on functionality to be as loose as possible while still satisfying the requirements of the system \cite{bennett2023b,bennett2022c}. Weakness, as it is called, is a measure of function as opposed to form. As such, maximising the weakness of constraints on function (w-maxing) is not mutually exclusive with simp-maxing. Both can take place, and optimising a finite set of representations to express the weakest possible collective constraint on functionality will \textit{cause} simple forms to express weak constraints \cite{bennett2024b}. 

Given an abstraction layer (a language), w-maxing involves identifying policies or hypotheses that are as non-specific or `weak' as possible, whilst still satisfying basic requirements. In experiments involving binary arithmetic, w-maxing alone yielded 110-500\% improvement in generalisation rate over simp-maxing alone. W-maxing also involves delegating control to lower levels of abstraction. This reflects the biological polycomputational architecture self-organisation \cite{bongard2023,Gershenson2025,fields2020,levin2024,sole2022}. Biological systems are comparatively more adaptable than artificial intelligence because biology distributes control and delegates it to lower levels of abstraction \cite{bennett2025a}. In computer science terms, this is like programming in C instead of Python, or on a field programmable gate array instead of C. More efficient, bespoke implementations are possible when adaptation extends down to smaller scales and lower levels of abstraction. Early stage examples of computing systems that delegate control in this manner include soft-robotics \cite{man2019} and self-organising systems of nano particles \cite{borghi2024_brainlikehw,paroli2023_receptron}. Because it does not separate software or hardware, it simultaneously optimises for both sample and energy efficiency \cite{bennett2025a}. But to optimise hardware as above, one must search an infinite space of embodiments. This is a process of trial and error, or optimisation through selection. It could be thought of like a biological self-organising system searching the morphospace during development and regeneration \cite{levin2024}. Because of the enactive frame, w-maxing has been used to explain causal reasoning, language and consciousness in terms that apply to both AI and biological self-organising systems \cite{bennett2023c,bennett2023d,bennettwelshciaunica2024,bennett2025f}.

\section{Conclusion}
Recent history has been dominated by scale-maxed approximation. I like to call this period The Embiggening. The rapid improvements suggest we were bottle-necked by compute and data. Now there are diminishing returns \cite{kaplan2020scaling}. Models are expensive. There are far more problems for which we have little data than problems for which we have a lot of training data. Reliable precision and energy efficiency are increasingly important considerations. Perhaps scaling is no longer the easiest way forward. Better opportunities lie in w-maxing and simp-maxing. Newer models are hybrids like o3, not pure approximations as when GPT-3 was released. There has also been a great deal of discussion around the economic potential of autonomous agents \cite{perrier2025positionstopactinglike}. This is where architectures like AERA, NARS and Hyperon stand to shine. Yes scale-maxed approximations dominate, but a fusion is required for an artificial scientist.
%
%
%
\bibliographystyle{splncs04}
\bibliography{main}

\end{document}